# Co-channel Interference Cancellation for Space-Time Coded OFDM Systems Using Adaptive Beamforming and Null Deepening

Raungrong Suleesathira

**Abstract**—Combined with space-time coding, the orthogonal frequency division multiplexing (OFDM) system explores space diversity. It is a potential scheme to offer spectral efficiency and robust high data rate transmissions over frequency-selective fading channel. However, space-time coding impairs the system ability to suppress interferences as the signals transmitted from two transmit antennas are superposed and interfered at the receiver antennas. In this paper, we developed an adaptive beamforming based on least mean squared error algorithm and null deepening to combat co-channel interference (CCI) for the space-time coded OFDM (STC-OFDM) system. To illustrate the performance of the presented approach, it is compared to the null steering beamformer which requires a prior knowledge of directions of arrival (DOAs). The structure of space-time decoders are preserved although there is the use of beamformers before decoding. By incorporating the proposed beamformer as a CCI canceller in the STC-OFDM systems, the performance improvement is achieved as shown in the simulation results.

———————————— ◆ ————————————

## 1 INTRODUCTION

Space-time coding [1], [2] is a coding technique designed for multiple antenna transmission. In addition to the diversity improvement on flat fading channels, it can achieve a coding gain with no bandwidth sacrifice. Combining multiple receive antennas, space-time coding can minimize the effects of multipath fading and achieve high channel capacities in the MIMO (multiple-input multiple-output) systems [3]. Recently, there has been an increasing interest in providing high data rate services such as video conferencing, multimedia internet access and wide area networks over wideband wireless channels. Consequently, the transmitted signals experience frequency-selective fading. Thus, the channel induces intersymbol interference (ISI) which makes code performance degrade. Without equalization, a promising approach to mitigate ISI is OFDM technique [4] which is used in various standards of wireless communication systems. The high rate data stream is demultiplexed into a large number of subchannels. The number must be chosen to ensure that each subchannel has a bandwidth less than the coherence bandwidth of the channel, so the subchannel is characterized as flat fading. Each subchannel however has a high error probability in deep fades.

The combination of space-time coding and the OFDM modulation [5] is not only alleviate drawbacks of each other, but also improve the high speed transmission performance limited to two fundamental impairments, multipath fading and ISI. However, using multiple transmit antennas at each mobile causes mutual interference at the receiver by either signals transmitted from different antennas of the same transmitter or other transmitters. Additional processing is required to mitigate co-channel interferences (CCI) in space-time coded OFDM (STC-OFDM) systems.

To maximize the STC-OFDM system efficiency, the problem of CCI introduced by the space-time coding must be solved. Adaptive antenna arrays are an attractive solution because they can suppress CCI and mitigate the effects of multipath fading [6], [7], [8], [9]. Regarding to beamforming techniques for STC-OFDM systems, some research attempts have been mainly focused on the transmit beamforming in downlink since the fact that download intensive services and wireless web browsing are to be introduced in the next generation [10], [11], [12], [13]. Study on receive beamforming which is widely applied to uplink of cellular mobile systems has also attracted attention to both suppress CCI and minimize fading effects. This latter is the case that we consider in this paper. In [14], it is shown that the scheme of MIMO wireless systems incorporating a beamforming method before space-time decoder can effectively mitigate CCI while preserving the space-time structure. The beamforming method called the minimum variance distortion response (MVDR) beamformer is used as a CCI canceller in OFDM systems using space-time trellis codes in reverse link. One disadvantage of the MVDR beamformer is based on accurate estimation of the desired DOAs and degrade obviously when there are errors in desired DOAs. To improve the robustness to uncertainty in DOAs of the MVDR method,

————————————————

• *Raungrong Suleesathira is with Department of Electronic and Telecommunication Engineering, Faculty of Engineering, King Mongkut's University of Technology Thonburi, Bangmod, Tungkru, Bangkok, 10140, Thailand.*

a robust beamforming algorithm based on particle filter is proposed in [15]. A minimum mean squared error (MMSE) beamforming is applied to the space-time block coding combined with the OFDM system [16]. The method relies on the number of training blocks corresponding to the chosen cluster sizes. Space-frequency OFDM system with the null steering beamforming for CCI suppression is investigated in [17]. In this paper, an adaptive beamforming algorithm based on minimum means square error of pilot signals is presented which does not require known desired DOAs as the MVDR beamforming and the null steering beamforming. The resulting coefficients are inversed in order to deepen nulls such that it can increase CCI suppression ability in the space-time trellis coded OFDM systems.

This paper is organized as follows. In section 2, the conventional STC-OFDM system model is given. Section 3 describes the proposed adaptive beamforming and null deepening as a CCI canceller in the STC-OFDM system. It also includes a conventional beamforming algorithm called the null steering beamforming. The space-time decoders are expressed in section 4. The simulation results are presented in section 5. Finally, conclusion is in section 6.

## 2 STC-OFDM MODEL

The conventional STC-OFDM scheme which includes arrays of $n_T$ transmit and $n_R$ receive antennas is illustrated in Fig.1. A transmitter shown in Fig. 1(a) begins by encoding a block of symbols to generate a space-time codeword. At time $t$, the space-time encoder constructs a matrix of $n_T \times K$ modulated symbols given as [1]

$$\mathbf{X}_t = \begin{bmatrix} x_{t,1}^1 & x_{t,2}^1 & \cdots & x_{t,K}^1 \\ x_{t,1}^2 & x_{t,2}^2 & \cdots & x_{t,K}^2 \\ \vdots & \vdots & \ddots & \vdots \\ x_{t,1}^{n_T} & x_{t,2}^{n_T} & \cdots & x_{t,K}^{n_T} \end{bmatrix}$$

where an element $x_{t,k}^i$ belongs to a constellation from M-ary phase-shift keying. The i-th row, $i = 1,\ldots,n_T$, represents a data sequence sent out the i-th transmit antenna. At the i-th transmit antenna, the serial-to-parallel converter allows us to obtain parallel data. OFDM modulation is used to modulate the parallel data, $x_{t,1}^i, x_{t,2}^i, \ldots, x_{t,K}^i$, by the IFFT. The total available bandwidth of $F$ Hz is divided into $K$ overlapping subcarries. In the time domain, a cyclic prefix (CP) is added to each OFDM frame during the guard time interval. The length of CP must be larger than the maximum time delay of multipath fading channel such that it can be avoided ISI due to the delay spread of the channel. Consider the receiver shown in Fig. 1(b). OFDM demodulation computes the FFT and removes CP. The output of the OFDM demodulator for the k-th subcarrier, $k = 1,\ldots,K$, and for the j-th receive antenna,

$j = 1,\ldots,n_R$, can be expressed in the frequency domain as

$$R_{t,k}^j = \sum_{i=1}^{n_T} H_{t,k}^{j,i} x_{t,k}^i + N_{t,k}^j \qquad (1)$$

where $H_{t,k}^{j,i}$ is the channel frequency response of the k-th subcarrier between the i-th transmit and j-th receive antennas in the presence of the noise sample $N_{t,k}^j$. The maximum likelihood decoding rule amounts to the computation of

$$\hat{\mathbf{X}}_t = \arg\min_{\hat{\mathbf{X}}} \sum_{j=1}^{n_R} \sum_{k=1}^{K} \left| R_{t,k}^j - \sum_{i=1}^{n_T} H_{t,k}^{j,i} x_{t,k}^i \right| \qquad (2)$$

The minimization is performed over all possible space-time codewords. Note that interleavers are optional to improve the performance.

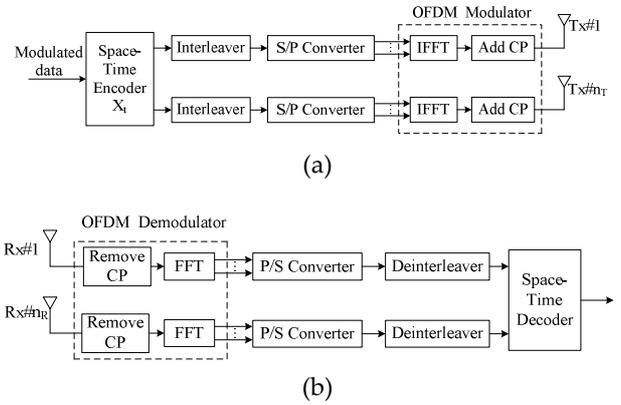

Fig. 1. Conventional $n_T \times n_R$ STC-OFDM system (a) STC-OFDM transmitter and (b) STC-OFDM receiver.

## 3 CCI CANCELLATION FOR STC-OFDM SYSTEM

In this section, we modified the STC-OFDM system to increase the capability to mitigate interferences. Fig. 2 and 3 present the block diagram of the proposed STC-OFDM system with CCI cancellation. Due to achievable diversity gain and coding gain, space-time Trellis codes (STTC) is chosen as a space-time encoder. In Fig. 2, the proposed transmitter multiplexes the STTC data and pilot symbols to generate a data sequence $\mathbf{x}_t^i$. By performing the IFFT on $\mathbf{x}_t^i$, the resulting signal $\mathbf{y}_t^i$ can be written in a vector form as [8]

$$\mathbf{y}_t^i = \mathbf{F}^H \mathbf{x}_t^i, \qquad i = 1,\ldots,n_T \qquad (3)$$

where $\mathbf{x}_t^i = \begin{bmatrix} x_{t,1}^i & x_{t,2}^i & \cdots & x_{t,K}^i \end{bmatrix}^T$,

$\mathbf{y}_t^i = \begin{bmatrix} y_{t,1}^i & y_{t,2}^i & \cdots & y_{t,K}^i \end{bmatrix}^T$ and

$$\mathbf{F} = \begin{bmatrix} 1 & 1 & \cdots & 1 \\ 1 & e^{-j2\pi(1)(1)/K} & \cdots & e^{-j2\pi(1)(K-1)/K} \\ \vdots & \vdots & \ddots & \vdots \\ 1 & e^{-j2\pi(K-1)(1)/K} & \cdots & e^{-j2\pi(K-1)(K-1)/K} \end{bmatrix}.$$

Matrix $\mathbf{F}$ is referred to as the FFT operation. Superscript $H$ denotes complex conjugate transposition of a matrix.

In the following discussion, a model consisting of one desired mobile and one interfering mobile is considered. Note that it can be extended for more than one presented CCI. Each mobile has two transmit antennas ($n_T = 2$). The base station has four receive antennas ($n_R = 4$). Assume that the signals propagate from the two transmit antennas of a mobile to the base station in the same DOA with two independent fading channels. Let's denote two independent paths in each DOA using vector notation as

$$\mathbf{h}_p = [h_p^1 \; h_p^2] \qquad p = 1,2 \qquad (4)$$

where $h_p^i$ is the $p$-th fading path of the $i$-th transmit antenna. After demodulation and CP removal from each frame, the received signal at the antenna array in Fig. 3 can be modelled as

$$\mathbf{V}_t = \sum_{p=1}^{2} \mathbf{a}(\theta_p)\mathbf{h}_p \mathbf{Y}_{t-\tau_p} + \sum_{p=1}^{2} \mathbf{a}(\theta_{p,\text{CCI}})\mathbf{h}_{p,\text{CCI}} \mathbf{Z}_{t-\tau_p} + \mathbf{N}_t. \quad (5)$$

The first term in (5) represents the signals received from the desired mobile while the second term represents the CCI. The signal transmitted from the desired mobile is given by

$$\mathbf{Y}_{t-\tau_p} = [\mathbf{y}_{t-\tau_p}^1, \mathbf{y}_{t-\tau_p}^2]^T \qquad (6)$$

where $\mathbf{y}_{t-\tau_p}^i$ is the signal transmitted from antenna $i$ at time $t(\mathbf{y}_t^i)$ that is propagating in fading path $p$ and has a time delay $\tau_p$. The antenna array response steering in the direction of arrival $\theta_p$ is defined as [18]

$$\mathbf{a}(\theta_p) = \left[1 \; e^{-j\pi \sin\theta_p}, \ldots, e^{-j\pi(n_R-1)\sin\theta_p}\right]^T. \quad (7)$$

In (5), the notation of the second term has the same meaning as the first term but it is for CCI. $\mathbf{N}_t$ is $n_R \times K$ an additive white Gaussian noise matrix with zero mean and variance $\sigma_n^2$. Its correlation matrix is $\mathbf{R} = E[\mathbf{N}_t \mathbf{N}_t^H] = \sigma_n^2 \mathbf{I}$.

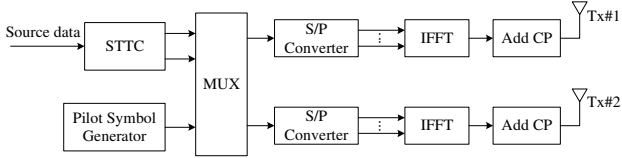

Fig. 2. Proposed STC-OFDM transmitter

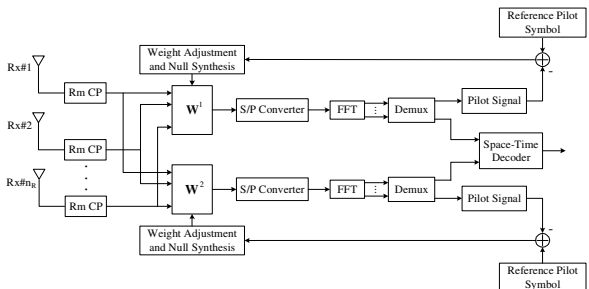

Fig. 3. STC-OFDM receiver with CCI canceller

To suppress CCI, an adaptive beamformer is introduced before the space-time decoder as seen in Fig. 3. Beamformer serves as a spatial filter to receive a strong signal from a particular direction while simultaneously blocking it from another direction. The beam response depends on the DOAs of the desired signal and DOAs of CCI impinging on the antenna array. Let $\mathbf{r}_t^l$ be the $l$-th beamformer output at time $t$ given by

$$\mathbf{r}_t^l = (\mathbf{w}_t^l)^H \mathbf{V}_t \qquad l = 1,2 \qquad (8)$$

where $\mathbf{w}_t^l = \left[w_{t,1}^l \; w_{t,2}^l, \ldots, w_{t,n_R}^l\right]^T$ is a weight vector which is crucial to form a beam response properly. After the FFT operation, the received signal vector in the frequency domain becomes

$$\begin{aligned}\tilde{\mathbf{y}}_t^l &= \mathbf{F}(\mathbf{r}_t^l)^H \\ &= \left[\tilde{y}_{t,1}^l \; \tilde{y}_{t,2}^l, \ldots, \tilde{y}_{t,K}^l\right]^T.\end{aligned} \qquad (9)$$

The signal is then demultiplexed to perform weight adapting and space-time decoding. In the following subsections, we presented an algorithm of adaptive beamforming and null deepening to control the weights. The conventional beamforming known as the null steering beamforming to combat CCI is also briefly explained. Additionally, expressions of the space-time decoder are also given to analyze the effect of CCI cancellation to the decoded codewords.

### 3.1 CCI Canceller using Adaptive Beamforming and Null Deepening

In this adaptive algorithm, the antenna weights are updated to minimize mean squared error. The error quantity is defined as the difference between the reference pilot symbols $y_{t,q}^l$ and the received pilot signal $\tilde{y}_{t,q}^l$. Accordingly, the mean square error (MSE) can be expressed as

$$E\left[(e_t^l)^2\right] = \sum_{q=1}^{Q} E\left(\left|y_{t,q}^l - \tilde{y}_{t,q}^l\right|^2\right) \qquad (10)$$

where $Q$ is the number of pilot symbols which are multiplexed to the STTC at the receiver. Let $\mathbf{y}_{t,Q}^l$ and $\tilde{\mathbf{y}}_{t,Q}^l$ be the reference pilot signals and the received pilot signals of $Q$ sub-carriers, respectively, given by

$$\begin{aligned}\mathbf{y}_t^l &= \left[y_{t,1}^l \; y_{t,2}^l, \ldots, y_{t,Q}^l\right]^T \\ \tilde{\mathbf{y}}_t^l &= \left[\tilde{y}_{t,1}^l \; \tilde{y}_{t,2}^l, \ldots, \tilde{y}_{t,Q}^l\right]^T.\end{aligned}$$

Using matrix notation, the error vector in frequency domain of (10) becomes

$$\mathbf{e}_t^l = \mathbf{y}_{t,Q}^l - \tilde{\mathbf{y}}_{t,Q}^l \qquad (11)$$

Transform the error vector into the time domain by

$$e_t^l = \mathbf{F}_Q^H \mathbf{e}_t^l$$
$$y_t^l = \mathbf{F}_Q^H \mathbf{y}_{t,Q}^l \quad (12)$$
$$\tilde{y}_t^l = \mathbf{F}_Q^H \tilde{\mathbf{y}}_{t,Q}^l$$

where

$$\mathbf{F}_Q = \begin{bmatrix} 1 & 1 & 1 & 1 \\ 1 & e^{-j2\pi(1)(1)/K} & \cdots & e^{-j2\pi(K-1)(1)/K} \\ \vdots & \vdots & \ddots & \vdots \\ 1 & e^{-j2\pi(1)(Q-1)/K} & \cdots & e^{-j2\pi(K-1)(Q-1)/K} \end{bmatrix}$$

Substituting (8) and (9) into $\tilde{y}_{t,Q}^l$ in (12) yields the error vector in the time domain as

$$e_t^l = y_t^l - (\mathbf{V}_t)^H \mathbf{w}_t^l. \quad (13)$$

Weight adjustment by the least mean squares (LMS) algorithm to obtain minimum errors is [7]

$$\mathbf{w}_{t+1}^l = \mathbf{w}_t^l - 2\mu \frac{\partial (e_t^l)^H}{\partial (\mathbf{w}_t^l)^*} e_t^l \quad (14)$$

where $\frac{\partial}{\partial \mathbf{w}^*}$ is the conjugate derivative with respect to the complex weight vector $\mathbf{w}$ and $\mu$ is a positive scalar that controls the convergence characteristic of the algorithm. Using the error vector in (13), the weight vector of the $l$-th beamformer at time $t$ is adjusted by the following recursive relation

$$\mathbf{w}_{t+1}^l = \mathbf{w}_t^l + 2\mu \mathbf{V}_t \mathbf{F}_Q^H \left( \mathbf{y}_{t,Q}^l - \tilde{\mathbf{y}}_{t,Q}^l \right) \quad (15)$$

Let $\mathbf{w}^l$ be the converged weight vector of the $l$-th beamformer. The beam response can be formed using the function expressed as

$$\mathbf{b}^l(\theta) = (\mathbf{w}^l)^H \mathbf{a}(\theta). \quad (16)$$

To increase the capability of interference cancellation, deep nulls are synthesized at the all CCI directions i.e. at $\theta_{CCI,1}, \theta_{CCI,2}$ and at the direction of one path $\theta_p$ where $p \neq l$ called "not in a look direction". Equation (16) can be rewritten in a matrix form as [19]

$$\begin{bmatrix} b^l(\theta_1) \\ b^l(\theta_2) \\ \vdots \\ b^l(\theta_N) \end{bmatrix} = \mathbf{D} \begin{bmatrix} (w_1^l)^* \\ (w_2^l)^* \\ \vdots \\ (w_{n_R}^l)^* \end{bmatrix} \quad (17)$$

where $N$ is the number of samples in the beam response and matrix $\mathbf{D}$ is

$$\mathbf{D} = \begin{bmatrix} 1 & e^{-j\pi \sin \theta_1} & \cdots & e^{-j\pi(n_R-1)\sin \theta_1} \\ 1 & e^{-j\pi \sin \theta_2} & \cdots & e^{-j\pi(n_R-1)\sin \theta_2} \\ \vdots & \vdots & \ddots & \vdots \\ 1 & e^{-j\pi \sin \theta_N} & \cdots & e^{-j\pi(n_R-1)\sin \theta_N} \end{bmatrix}.$$

The procedure of placing a null at any direction $\theta_d$ with an angle width of $\Delta \theta_d$ is carried out by the following steps.

1. Impose a null on the response $\mathbf{b}^l(\theta)$ by replacing $M$ samples with zeros. The chosen samples are located in the range of $\Delta \theta_d$ around $\theta_d$. Let $\mathbf{b}_{null}^l(\theta)$ be the beam response with the imposed null given by

$$b_{null}^l(\theta_n) = \begin{cases} 0 & (\theta_d - \frac{\Delta \theta_d}{2}) \leq \theta_n \leq (\theta_d + \frac{\Delta \theta_d}{2}) \\ b^l(\theta_n) & \text{elsewhere} \end{cases}$$
$$\text{for} \quad n = 1, \ldots, N$$

2. Apply the inverse beam response to recalculate the weights of the response constrained with the prescribed nulls by $\mathbf{w}_{null}^l = \mathbf{D}^{-1} \mathbf{b}_{null}^l(\theta)$.

3. Substitute the weight vector $\mathbf{w}_{null}^l$ into (16) to achieve deep null synthesis.

In the next subsection, the null steering beamformer which can perfectly steer the main beam in the desired direction and impose nulls in the directions of CCI signal will be briefly reviewed.

### 3.2 CCI Canceller using Null Steering Beamforming

In the null steering beamforming, the desired directions of a mainbeam and the direction of a null must be prior known to generate the weight vector. For the $l$-th beamformer in Fig. 3, let $\mathbf{a}(\theta_l)$ be the steering vector where the unity response is required and $\mathbf{a}(\theta_{CCI,l})$ be the steering vector for nulling. Then, the weight vector of the $l$-th beamformer satisfies [18]

$$(\mathbf{w}^l)^H \mathbf{A} = (\mathbf{c}^l)^T, \quad l = 1, 2 \quad (18)$$

where $\mathbf{A} = [\mathbf{a}(\theta_1) \; \mathbf{a}(\theta_2) \; \mathbf{a}(\theta_{CCI,1}) \; \mathbf{a}(\theta_{CCI,2})]$ with columns being the steering vectors associated with all sources. The constraint for each beamformer is $\mathbf{c}^1 = [1 \; 0 \; 0 \; 0]^T$ when $l = 1$ and $\mathbf{c}^2 = [0 \; 1 \; 0 \; 0]^T$ when $l = 2$. The solution for the weight vector is given by

$$(\mathbf{w}^l)^T = (\mathbf{c}^l)^T \mathbf{A}^{-1}. \quad (19)$$

## 4 SPACE-TIME DECODER

For STTC, the Viterbi algorithm [20] is employed to perform the maximum likelihood decoding. Assume that channel state information is available at the receiver. The Viterbi algorithm selects the path with the minimum distance metric as the decoded sequence. The derivations of branch metric corresponding to the previously mentioned two beamformers are discussed in the following subsections.

### 4.1 Space-time Decoder after Adaptive Beamforming and Null Deepening

Using the response of the adaptive beamforming and null deepening, the beamformer outputs are affected by the two paths from the desired mobile and the residual CCI. For simplicity, we assume that the residual CCI is negligible and the mainbeam has a unity response. Substituting (5) into (8), the $l$-th beamformer output can be expressed as

$$\begin{aligned}\mathbf{r}_t^l &= (\mathbf{w}_t^l)^H \sum_{p=1}^{2} \mathbf{a}(\theta_p)\mathbf{h}_p \mathbf{Y}_{t-\tau_p} + (\mathbf{w}_t^l)^H \mathbf{N}_t \\ &= (\mathbf{w}_t^l)^H \mathbf{a}(\theta_1)\mathbf{h}_1 \mathbf{Y}_{t-\tau_1} + (\mathbf{w}_t^l)^H \mathbf{a}(\theta_2)\mathbf{h}_2 \mathbf{Y}_{t-\tau_2} \\ &+ (\mathbf{w}_t^l)^H \mathbf{N}_t.\end{aligned} \quad (20)$$

Assuming that $\tau_1 = 0$ and $\tau_2 = \tau$, thus we can omit the subscript $\tau_1$. Equation (20) becomes

$$\mathbf{r}_t^l = (\mathbf{W}_t^l)^H \mathbf{a}(\theta_1)\mathbf{h}_1 \mathbf{Y}_t + (\mathbf{w}_t^l)^H \mathbf{a}(\theta_2)\mathbf{h}_2 \mathbf{Y}_{t-\tau} + (\mathbf{w}_t^l)^H \mathbf{N}_t \quad (21)$$

The output $\mathbf{r}_t^l$ is transformed into frequency domain at the $k$-th subcarrier as

$$R_{t,k}^l = \sum_{i=1}^{2} H_{t,k}^{l,i} x_{t,k}^i + u_{t,k}^l \quad l=1,2 \quad (22)$$

where $u_{t,k}^l$ is the Fourier transform of $(\mathbf{w}_t^l)^H \mathbf{N}_t$. The frequency channel response is

$$\begin{bmatrix} H_{t,k}^{1,1} & H_{t,k}^{1,2} \\ H_{t,k}^{2,1} & H_{t,k}^{2,2} \end{bmatrix} = \begin{bmatrix} h_1^1 + \rho h_2^1 \omega & h_1^2 + \rho h_2^2 \omega \\ \beta h_1^1 + h_2^1 \omega & \beta h_1^2 + h_2^2 \omega \end{bmatrix} \quad (23)$$

where $\rho = (\mathbf{w}_t^1)^H \mathbf{a}(\theta_2)$, $\beta = (\mathbf{w}_t^2)^H \mathbf{a}(\theta_1)$ and $\omega = e^{-j2\pi\tau k/K}$. The space-time decoder using the Viterbi algorithm [20] to obtain decoded symbols become

$$\hat{\mathbf{x}}_t = \arg\min_{\hat{\mathbf{x}}} \sum_{l=1}^{2} \sum_{k=1}^{K} \left( R_{t,k}^l - \sum_{i=1}^{2} H_{t,k}^{l,i} x_{t,k}^i \right)^* \left( (\mathbf{w}_t^l)^H \mathbf{w}_t^l \right)^{-1} \left( R_{t,k}^l - \sum_{i=1}^{2} H_{t,k}^{l,i} x_{t,k}^i \right) \quad (24)$$

### 4.2 Space-time Decoder after Null Steering Beamforming

Using the response of the null steering beamformer, all CCI signals except the desired signal from one path in a look direction can be completely cancelled. Substituting (19) into (8), the $j$-th beamformer output can be expressed as

$$\mathbf{r}_t^l = \mathbf{h}_l \mathbf{Y}_{t-\tau_l} + (\mathbf{w}_t^l)^H \mathbf{N}_t \quad (25)$$

This can be viewed as time diversity with the frequency channel response in (22) given as

$$\begin{bmatrix} H_{t,k}^{1,1} & H_{t,k}^{1,2} \\ H_{t,k}^{2,1} & H_{t,k}^{2,2} \end{bmatrix} = \begin{bmatrix} h_1^1 & h_1^2 \\ h_2^1 \omega & h_2^2 \omega \end{bmatrix}. \quad (26)$$

Implement Viterbi algorithm to find the branch metric using (24) with the frequency response of (26).

## 5 SIMULATION RESULTS

In this section, simulations are conducted to evaluate the performance of the proposed adaptive beamforming and null deepening as a CCI canceller in the STC-OFDM system. We use a 16-state space-time trellis coded QPSK scheme. The encoder takes 512 bits as its input at each time to generate two coded QPSK sequences of length 256 symbols which are multiplexed to 32 pilot symbols. The transmitted signals experience the frequency selective, two-ray Rayleigh fading channel with a delay of $\tau = 15 \ \mu s$. Both two paths of a transmitted signal have equal average power. In fading channel with a maximum Doppler frequency of 50 Hz, the power spectral density satisfies Jakes model [21]. The total available bandwidth is 1 MHz with 288 subcarriers. This corresponds to a subchannel separation of 3.5 KHz and the duration of an OFDM frame is 288 μs. To avoid ISI, a CP of 40 μs duration is appended to each OFDM frame. In the following simulation, the signal to noise ratio (SNR) is defined as the average energy per symbol divided by the one sided noise power spectral density $N_0$, and the signal to interference ratio (SIR) is the ratio between average energy per symbol of the desired transmitter and CCI.

Figs. 4 and 5 illustrate the beam responses generated by the proposed adaptive beamforming and null deepening in comparison to the null steering beamforming algorithm. The beam patterns formed by the first beamformer are shown in Fig. 4 while the beam patterns formed by the second beamformer are shown in Fig. 5. The impinging angles of desired mobile and the CCI are $\theta_1$ = 10, $\theta_2$ = -20, $\theta_{CCI,1}$ = -60, $\theta_{CCI,2}$ = 40. The null width is fixed at $\Delta\theta_d = 5$ to deepen nulls at the directions of CCI and an unlook direction of the desired user. For each beam response, the CCI interferers and unlook direction of the desired signal can be suppressed significantly. More specifically, the main beam in Fig. 4 is to steer the desired signal at $\theta_1$ = 10 and nulls at -20, -60, and 40 while the main beam in Fig. 5 is to steer the desired signal at $\theta_2 = 20$ and nulls at 10, -60 and 40. It can be seen that we can achieve deeper nulls by using the proposed null synthesis. This reduces interference signals significantly. Note that the proposed beamformer does not require the prior knowledge of the DOAs as the null steering beamformer does.

In Fig. 6, we present the performance of the conventional STC-OFDM system without CCI cancellers with different SIR values. As SIR increases, the performance improves according to the decreased frame error rates (FER). This implies that CCI degrades the performance of the STC-OFDM system.

In addition to the presence of CCI signals, duration of delay spread also affects the performance degradation of the conventional STC-OFDM scheme as illustrated in Fig. 7. Due to the fact that multipath channels with a large delay spread give rise to a short coherence bandwidth, this yields a much higher frequency diversity gain for the STTC than that of short delay spread. The time delay has no impact when applying the proposed beamformer as a CCI canceller.

Fig. 8 indicates that performance of proposed system is not affected by a time delay because the guard interval is larger than the time delay. Therefore, the beamformers serve as a CCI canceller as well as a time diversity technique.

We compare the performance of the STC-OFDM system using the adaptive beamforming and null deepening for the different number of interferers as plotted in Fig. 9. The DOAs of the two paths of the desired mobile are (10, -20) and the DOAs of the two paths of one interferer are given at (-60, 40). For two interferes, we use (20, -60) for the angles of the two paths of the desired mobile whereas (50, -20, 80, 0) represents the DOAs of the two paths of two interferes. Let (20, -60, 50, -20, 80, 0, 35, -45) be the third DOA set where first two angles are the DOAs of the desired mobile and the last six angles are the DOAs of the three interferes. As seen in Fig. 9, it performs worse as the number of interferes increases.

Fig. 10 plots FERs to compare the performances of the STC-OFDM scheme between using the proposed beamformer (solid lines) and the null steering beamformer (dash lines). Three different impinging directions of the two paths of the desired signals are considered. The performance of the STC-OFDM scheme using the null steering beamformer is affected by the impinging angles. For the STC-OFDM scheme using the null steering beamformer, the FERs (dash line) increase if DOAs is closer. The null steering beamforming gives unsharp beam patterns due to the noise gain induced when two impinging angles of the desired signals become close to each other.

We conclude the ability to suppress CCI of the proposed beamformer by presenting Fig. 11. The performance of the conventional STC-OFDM scheme is used as an upper bound to show the system improvement using the proposed beamformer. The performance of the proposed adaptive scheme based on the MMSE algorithm and null deepening outperforms the method based on the null steering beamforming.

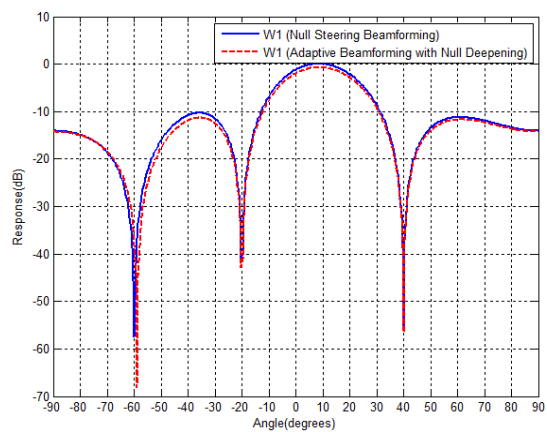

Fig. 4. Beam responses using $\mathbf{w}^1$ generated by the adaptive beamforming and null deepening in comparison to the null steering beamforming

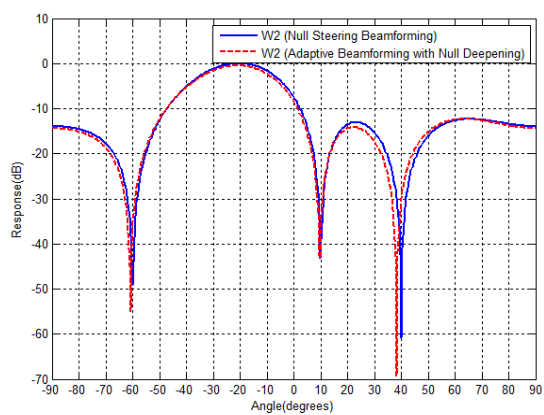

Fig. 5. Beam responses using $\mathbf{w}^2$ generated by the adaptive beamforming and null deepening in comparison to the null steering beamforming

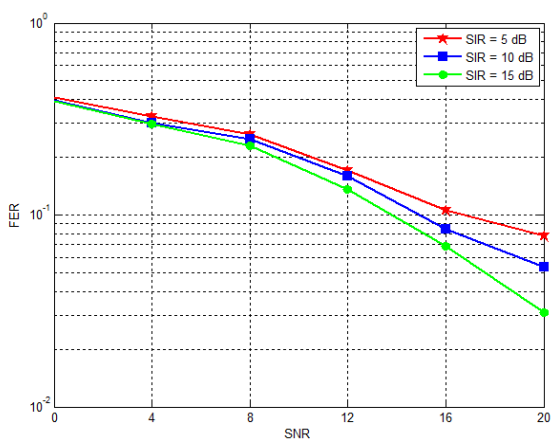

Fig. 6. FER performance of the conventional STC-OFDM system for different SIRs

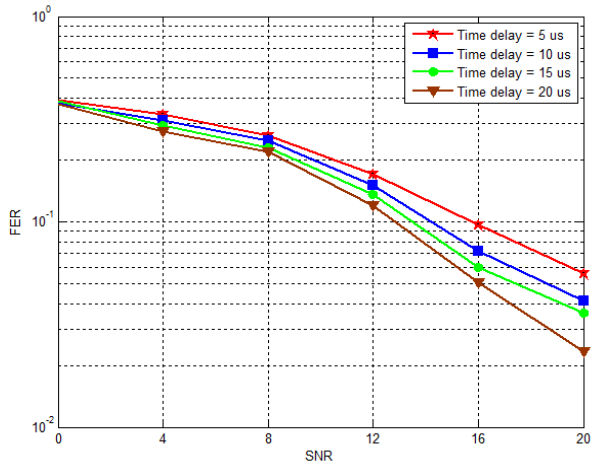

Fig. 7. FER performance of the conventional STC-OFDM system for different time delays

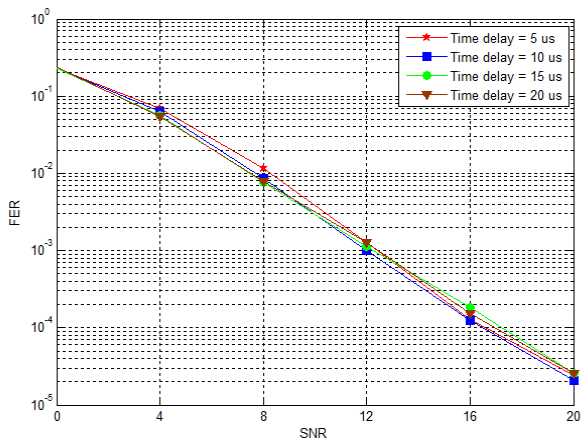

Fig. 8. FER performance of the proposed STC-OFDM system for different time delays

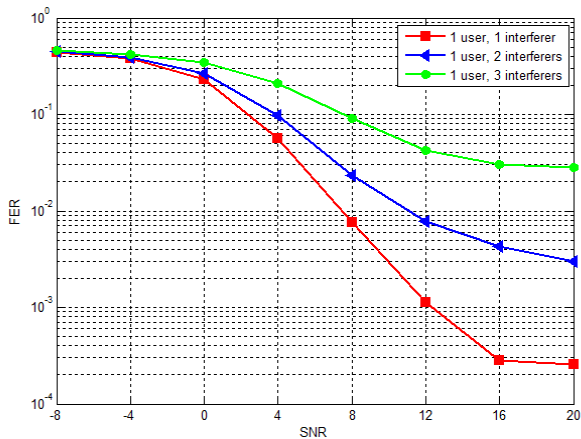

Fig. 9. FER performance of the proposed STC-OFDM system for different number of interferers

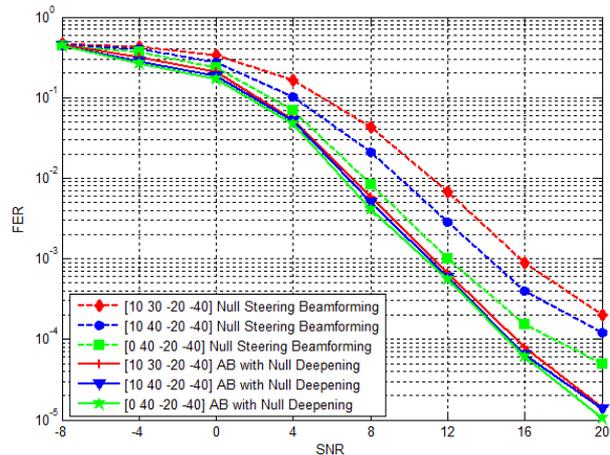

Fig. 10. Performance comparison between the adaptive beamforming (AB) and null deepening and the null steering beamforming for different impinging angles of the two paths of the desired signals

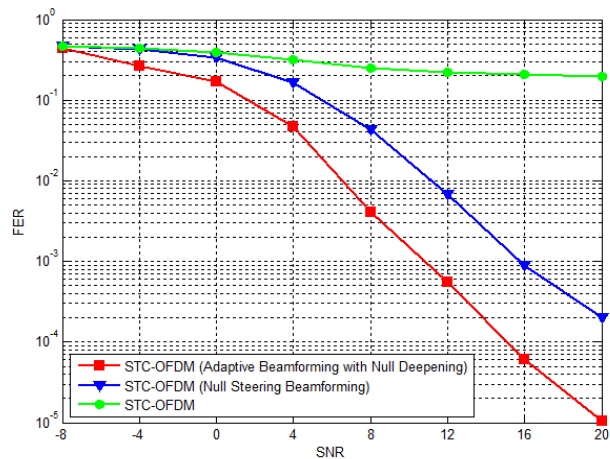

Fig. 11. FER comparison of the STC-OFDM using the adaptive beamforming and null deepening and the null steering beamforming

## 6 CONCLUSION

CCI canceller using the LMS adaptive beamformer and null deepening is presented in the STC-OFDM system. Due to time diversity of multipath propagation, the proposed beamformer can significantly suppress CCI for each path and preserve the structure of space-time decoding. The null steering beamformer which requires a prior knowledge of DOAs is used to compare the performance of the proposed CCI canceller. It is demonstrated that the proposed algorithm outperforms the null steering beamformer to increase the system ability for CCI cancellation.

## ACKNOWLEDGMENT

This work was supported in part by the Thailand Research Fund (TRF) under Grant MRG5080005.

**Raungrong Suleesathira** received the B.S. degree from the Kasetsart University, Bangkok, Thailand, in 1994, and the M.S. and Ph.D. degrees from the University of Pittsburgh, Pa, USA, in 1996 and 2001, respectively, all in electrical engineering. Since 2001, she has been on the faculty of the Department of Electronic and Telecommunication Engineering at King Mongkut's University of Technology Thonburi, Bangkok, Thailand, where she is now an Associate Professor. She was an assistant editor of ECTI Transactions on Electrical Engineering, Electronics, and Communications during the year of 2005-2008. Her current research interests are in the area of wireless communication and cognitive radio.